# A System for Extracting Sentiment from Large-Scale Arabic Social Data

Hao Wang, Vijay R. Bommireddipalli
Silicon Valley Lab
IBM
San Jose, USA
Email: {haowang, vijayrb}@us.ibm.com

Ayman Hanafy, Mohamed Bahgat, Sara Noeman,
Ossama S. Emam
Cairo Human Language Technologies Group
IBM
Cairo, Egypt
Email: {ahanafy, mohamedb@ie, noemans@eg, emam@eg}.ibm.com

*Abstract*—Social media data in Arabic language is becoming more and more abundant. It is a consensus that valuable information lies in social media data. Mining this data and making the process easier are gaining momentum in the industries. This paper describes an enterprise system we developed for extracting sentiment from large volumes of social data in Arabic dialects. First, we give an overview of the Big Data system for information extraction from multilingual social data from a variety of sources. Then, we focus on the Arabic sentiment analysis capability that was built on top of the system including normalizing written Arabic dialects, building sentiment lexicons, sentiment classification, and performance evaluation. Lastly, we demonstrate the value of enriching sentiment results with user profiles in understanding sentiments of a specific user group.

## I. INTRODUCTION

Enterprises have abundant structured and unstructured data from both internal and external sources, which is often analyzed and mined using computers and algorithms in order to provide better products and services to their customers. With the rise of popularity of social media in the last decade and the availability of social data, enterprises now have a quickly growing interest in getting access to and analyzing social media data to enhance the value of their internal data and gain competitive advantage over competitors. Sentiment analysis is particularly interesting and useful for that purpose. For example, a retailer can use polarity in sentiment to determine customers' preferences and opinions on a product; a financial service company can track customers' feedback on social media or from surveys to improve its services. Since we were frequently seeing requirements of social data analytics from the industry, we started building systems to facilitate the analysis of social media data. Such requirements were echoed in Arabic-speaking countries and led to the work described in this paper.

Arabic-speaking online population was growing tremendously in recent years. Arabic was ranked the 4th language in terms of the number of users on the Internet [1, 2]. Among all the online activities, social media was a top catalyst to the growth and has received a lot of attention and popularity in many Arabic-speaking countries fueled by recent events. For example, the proportion of Arabic posts on Twitter has grown from almost nothing in 2011 to the 6th among all the languages in 2013 [3].

Arabic is a particularly interesting language. It has been under extensive research in the domain of natural language processing with recent interests in social data analysis and typically sentiment analysis. Microblog posts in Arabic are abundant in Twitter traffic, one of the most popular social media in the world. Arabic has significant differences in terms of lexical and rich morphological features compared to, for example, Latin derived languages. Therefore, it poses special problems to natural language processing and sentiment analysis. One aspect of the differences is sentence structure. An Arabic sentence can be in the form of Verb-Subject-Object or Subject-Verb-Object. Sentences can also be constructed without a verb. Another aspect is the rich morphology in word form. A word can have multiple affixes and take many different forms.

Another challenge is from the domain of social media. Most written text in that domain is in different Arabic dialects and not the Modern Standard Arabic (MSA). Dialects in particular present new form of word inflections and sentence structure. The biggest challenge in that domain is the fact that there are no exact rules governing such derived dialect and can vary significantly even within the same country. In this work, we attempt to address this issue in Egyptian and Saudi dialects.

In the context of above-mentioned challenges, we built a system to extract sentiment and other kinds of information from large volumes of social data. Being able to process a large amount of data efficiently and quickly is key for acquiring accurate and current information. The size of data helps normalize statistics making it robust against noise while on the other hand it supports the identification of anomalies. Such analysis needs to be performed quickly to allow users to take the right actions at the right time that is a key aspect in a highly competitive environment and volatile customer base.

The underlying system used as our framework includes a variety of analytic capabilities, such as sentiment analysis, extracting demographic and geographic information and building user profiles. Because of the limited space, we will briefly describe the overall architecture of the system and focus on sentiment analysis for Arabic.

In this work we use a rule-based approach to perform sentiment analysis. This approach basically relies on: 1) a rich set of rules that can capture patterns expressing sentiment in the input; 2) a rich set of sentiment lexicons that can typically capture sentiment aspects in its different forms of expressions building. Many sentiment lexicon resources exist for English. But for Arabic and for a particular domain, the sentiment lexicon has to be built from scratch or enhanced from an existing one.

Multiple sources of social data are available. Here we focus on Twitter data as the source of social media data.

## II. RELATED WORK

In the past years, sentiment analysis and opinion mining are of keen interests to researchers and commercial companies partly because of its potential applications in many areas including marketing, politics, and communications research. Research on sentiment analysis has been conducted on a variety of genres of text (such as blogs, surveys, micro-blogs, product and movie reviews) using rule-based, machine learning or hybrid techniques [4, 5]. There were also efforts to build resources such as sentiment lexicon in different languages to support the task of classifying sentiment. Social media data is quickly becoming a rich source of opinionated text. For that reason, it is not surprising that sentiment in social media is now a popular subject of study [6-8].

However, most of sentiment analysis research focuses on English and other Indo-European languages. Little research in this area has been done for Arabic and its dialects [2, 9]. In particular, sentiment in Arabic social media received little attention from researchers although Arabic social media is gaining popularity in great momentum [9]. SAMAR [10] is a system to address the need of sentiment analysis in Arabic. There were a few more studies on this topic [11-14] and on sentence-level Arabic sentiment analysis [15]. To facilitate sentiment analysis in Arabic, there have been some efforts to build sentiment lexicon for Arabic either from translation of English sentiment lexicon or from scratch [16]. Annotated corpora [17, 18] were also created for Arabic sentiment analysis.

Despite these efforts, there still large need for contributions to Arabic sentiment analysis and research in the particular challenges posed by Arabic. We attempt to build a system to address many of the challenges and that is easily customizable and adaptable to new domains.

## III. THE SYSTEM

As stated earlier robust insights and results from social media require processing significant amount of data. Speed is also a requirement to help decision makers to react quickly to changes highlighted by analytics insights. Thus, in this section we describe the software platform that our sentiment analytics is developed upon.

The underlying platform, IBM BigInsights, is an Apache Hadoop-based product for Enterprise to manage and analyze large volumes of structured and unstructured data using commodity hardware. It includes many additional enterprise-grade features, such as data visualization and exploration, security and administration, text analytics and machine learning. It provides a web-based UI for common tasks such as managing the Hadoop cluster, managing files stored on the cluster, and running applications to process and analyze data.

Data from social media, blogs and online forums contains immerse and valuable information about customer behaviors and user profiles. However, loading and analyzing social data is not an easy problem due to the large volume, high velocity and variety of the data. Mining relevant information in social data requires special technical skills in many aspects, such as information extraction, sentiment analysis and entity resolution. To facilitate this task, a software product called Social Data Analytics (SDA) was created on top of the above-mentioned big data platform to provide the capabilities to ingest social data and extract relevant information from social data and build user profiles for specific use cases and industries with minimal effort.

With SDA, even a novice user can ingest and analyze social media data, identifying user characteristics like gender, locations, names, and hobbies; develop comprehensive user profiles across messages and sources; associate profiles with expressions of sentiment, buzz, intent, and ownership around brands, products, and companies. It enables data analysts with little knowledge in information extraction and sentiment analysis to get results from social data quickly. Currently it only supports English. This work is a part of the effort to add support for Arabic.

In SDA, a component called SystemT is used to build extractors to extract all the above-mentioned information. SystemT [19, 20] is an information extraction system built to make the process of developing extractors easier by providing specialized development tools. A core concept in SystemT is AQL [21], a declarative rule language with a SQL-like syntax. AQL replaces multiple obscure languages typically used to build extractors. Because AQL is a declarative language, rule developers can focus on what to extract, allowing SystemT's cost-based optimizer to determine the most efficient execution plan for the extractors. The sentiment extractor for Arabic described in this work is developed based on SystemT and using AQL. The choice of using a rule-based information extraction system instead of a machine learning approach was determined by our business and technical requirements: the software system needs to be adapted to different domains with expert knowledge from the particular domain; it needs to be developed, enhanced and maintained relatively easily; it also needs to provide customization points for users to fix errors and enhance the extraction results. With these considerations, a rule-based approach was a better option for us (see [22] for more discussion on this topic). Section 6.2 provides details on the AQL rules for the Arabic sentiment extractor.

## IV. NORMALIZATION OF ARABIC DIALECT TEXT

Majority of social media text often contains non-standard forms of writing. Such lack of standards can impact significantly the quality of information extracted from processed data. The social media content we are interested in is from Egypt and the Kingdom of Saudi Arabia. The posts generally are written in their local dialects that have no formal or standard method of writing. The same word may be written in different spelling variations and using different character sets. In such cases, coverage is negatively impacted significantly causing fewer cases to be captured and analyzed. To reduce the negative impact of non-standard text, the text is normalized at both character and word levels in the beginning of our analysis pipeline.

Arabic character level normalization is a common technique and widely used in Arabic NLP especially in social media retrieval tasks [16, 23]. Words are normalized into some forms that are considered as standard for the application in question. For example, three sets of Arabic characters are commonly confused in informal text: Alef-Hamza, Yeh-Alef Maksora, and Heh-Teh Marbota. In the first case, the word "أنت" (you) with an Alef-Hamza as the standard form in Modern Standard Arabic, is commonly written in dialects as "انت" with no Hamza above Alef, or incorrectly written as "إنت" with Hamza under Alef. The different forms of Alef are normalized to a single form: in our case the most common one (bare Alef). This form is considered the standard one in our system. Another example of normalization is changing Urdu style Heh (Unicode: 0x06C1) and Teh (Unicode: 0x06C3) which are common in Saudi informal writings to their equivalent in Arabic.

Some special characters are removed if they are not frequently used or if their use may not be always in the correct form. For example, Diacritics (Tashkeel; Unicode 0x064B to 0x065F) and Kasheeda (Unicode 0x0640) are completely removed from text. Moreover, a common practice in informal text is to repeat characters to show emphasis. In such cases repeated characters are removed to match a known word.

Another common practice is the use of Roman (ASCII) characters instead of Arabic characters to write Arabic. An example is the word "e7na" which should be converted to "احنا" (we). We've developed a phrase-based machine translation [24] like model but on character level to convert from Roman characters to Arabic. It takes into account different possible spellings in Roman where all variations are converted into a single frequent dialect or standard Arabic form. On social media, it is extremely common to write mixed Arabic in Roman characters and actual English words so our sentiment analysis provides a customizable exclusion list for words to be excluded from such kind of normalization.

## V. Building Arabic Sentiment Lexicons

Our approach to sentiment classification requires the creation of sentiment lexicons in positive and negative polarities. To build such resources we started with translating existing English sentiment resources in the current system to Standard Arabic (translating sentiment lexicons has been shown to be quite effective [16]). Then our linguists validated the lexicons and removed irrelevant cases.

The Standard Arabic lexicons serve as a common ground for creating sentiment lexicons in other dialects. The next step was to add Egyptian and/or Saudi colloquial dialect forms expressing different sentiments. For each Arabic word in the lexicon, the linguists added the Egyptian and Saudi equivalent dialect ones if exist in a different form in a kind of a semantic translation. For example, the Arabic expression "جميل جدا" (very beautiful) has an equivalent Egyptian one "حلو اوي" and Saudi one "حلو مرة" as well. Lexicons were enriched also by adding sentiment expressions and idioms. Also, it is common to express sentiment through conveying prayers or supplications either for in case of positive sentiment or against in case of negative sentiment. An example for negative sentiment expressions is the Arabic equivalent for "Burn in Hell" which depicts anger at someone. On the other hand the idiom "You are an angle" has positive sentiment and is being used when someone is in support or in favor. Since Arabic dialects differ from MSA phonologically and lexically and do not have standard orthography, the accuracies of morphological analyzers available for Arabic dialects are not comparable to that obtained for MSA (which already by itself has its own morphological complexity). Hence, Arabic morphological analyzers were not considered at this phase, and lexicons were enriched by adding most frequent forms for each word to increase the coverage. For example, for the positive word جميلة (beautiful), its frequent forms as "وجميلة", "الجميلة" have been added to the positive lexicon. The linguists enriched the lexicons with negation terms as well to handle the cases where the sentiment is reversed due to negation.

Disambiguation is another challenge. Some expressions overlap with sentiment terms. For example the expression "صباح الخير" (good morning) overlaps with the positive word "الخير" (good). Also, some male and female names are derived from adjectives that are confused for sentiment. Examples of female names are "لطيفة" (nice) and "جميلة" (beautiful), and male names are "حبيب" (lover) and "سعيد" (happy). Thus it was required to create a blocker lexicon (i.e., an exclusion list), which contains all words, expressions and person names that should not be classified as sentiment terms. To further enrich our polarity and blocker lexicons, the linguists checked a few thousands of Arabic tweets to find terms that were missing. They also enriched the lexicons by adding domain specific sentiment words and expressions to enhance the results for each of the use cases. We ended up having Egyptian and Saudi lexicons of about 14K and 18K entries respectively. Details about Lexicons sizes are shown in Table I. There are common MSA entries between Egyptian and Saudi lexicons. Also Saudi lexicons overlaps with Egyptian ones as some Egyptian terms are commonly used in Saudi dialect as well.

TABLE I. SIZES OF EGYPTIAN AND SAUDI DIALECT LEXICONS

| Dialect type | Positive | Negative | Total |
|---|---|---|---|
| Egyptian | 4411 | 9584 | 13995 |
| Saudi | 6844 | 11273 | 18117 |

## VI. SENTIMENT CLASSIFICATION

### A. Data set

In this paper, we focus on Twitter data as the source of social media content, although the system can take as input multiple social sources including Twitter, blogs and discussion boards. To measure the accuracy and performance of the system, Twitter data annotated with sentiment polarity is required. Some annotated Twitter data sets from past studies were made publicly available [15, 16]. However, we have to create a new annotated data set because the system will be initially applied to three use cases: Telecommunications in Egyptian Arabic (Egyptian Teleco), Government in Egyptian Arabic (Egyptian Government), and Employment in Saudi Arabic (Saudi Employment). To build the data set, Gnip Twitter API was used to collect Arabic tweets containing keywords for products and companies in Teleco, political events, jobs and employment. Table II shows some examples of the keywords we used for the Teleco use case.

The Twitter data set was collected based on several dimensions. The first dimension was the location from which the tweets originated. We gathered tweets mainly from countries speaking our target dialects (Egyptian Arabic and Saudi Arabic) and also included some Arabic tweets from other Arabic countries and around the world. The second dimension was the time of the tweets posted. For the Egyptian Government use case, we collected historical tweets that were posted during different events in the past. The last dimension was the language variations in the tweets. We tried to capture as much spelling variations as we can for each dialect and use case.

Duplicated tweets and re-tweets were removed. The final data set contains 5,000 tweets per use case (15,000 in total). It was then given to our Arabic annotators, who were asked to manually assign one of the three labels to each tweet: positive, negative and neutral. A tweet is discarded if the annotators did not agree on its classification. For polarity classification experiments, 400 tweets were randomly selected for each polarity category for every use case to create a balanced test set. Thus, the test set contains 1,200 tweets in total for each use case. The rest of the tweets were used as a development set to enrich sentiment lexicons as mentioned in section 5 as well as to develop and enhance sentiment rules. Sometimes a word could indicate completely different polarities in two use cases. Our sentiment rules and lexicons are created and tuned for each use case. Therefore, they are domain-specific and dialect-specific, although a large part of them is common among the use cases.

Besides the annotated data set we created to measure accuracy of the system, another Twitter data set is prepared to test the system's performance in analyzing large-scale data set. It includes a month of tweets collected through the Twitter Decahose API (a 10% sample of entire Twitter traffic) in January 2015, which contains about 1.05 billion tweets. With this large data set, we will show that the system adds value to sentiment analysis by enriching sentiment classification results with demographic and geographic information. For example, the sentiment results can be further broken down by the gender and location of the authors.

TABLE II. RELEVANT KEYWORDS OF PRODUCTS FOR THE TELECO USE CASE

| Keyword category | Arabic translation | English translation |
|---|---|---|
| Internet services | يو اس بي | USB |
| | اي دي اس ال | ADSL |
| Bundles | باقة اي فون | IPhone bundle |
| | باقة البلاك بيري | Black Perry bundle |
| Services | تحويل الرصيد | Credit transfer |
| | الفيديو كول | Video call |

### B. Analysis Sequence

Our sentiment analytic pipeline starts with normalizing the text (see Section 4) in each tweet to reduce data sparseness and to prepare data for the next step. The system then identifies relevant tweets by using a keyword-matching algorithm. All the relevant tweets are then assigned a sentiment label using a complex set of rules as well as polarity and blocker lexicons. The rules implemented in our system for Arabic are described below in three processing phases.

*1) Extraction phase:*

1. Extract polarity terms (words/phrases) from a tweet using polarity lexicons.
2. Extract blocker terms from a tweet using blockers lexicon.
3. Extract question marks and question words from a tweet as it is assumed that questions do not carry any sentiment.
4. Extract terms indicating a contrast in context (e.g., but and however in English) using a change-context lexicon. When a change-context term occurred, the polarity before the change-context term usually is the opposite of the polarity after it. For example, in
"عملكم رائع جدا وبرنامج متكامل ومفيد، لكن ينقصه شي وهو عدم توفير خيار بتغير الخط الي خط جهاز الجوال نفسه",
the word "لكن" (but) is the change-context term. The polarity before it is positive and the actual polarity after it is negative.

*2) Exclusion phase:*

1. Exclude extracted polarity terms that overlap with the extracted blocker term. For example, the extracted polarity term "الخير" will be discarded if it overlaps with the extracted blockers term "صباح الخير".
2. Exclude polarity terms if a tweet contains questions as determined by step 2 in Extraction phase. As an exception, terms having strong polarity are not excluded. For example, the polarity term "إيه القرف ده؟" (What a shet?), although it comes in question form, it is considered a strong polarity term and hence is not excluded. We created a small lexicon for strong polarity terms.

*3) Linkage phase:*

In this phase, we try to link through set of rules, the detected polarity terms to the detected objects-of-interest (e.g., a product name) within the tweet. Examples of such rules are:

1. A link is created if the assigned polarity and the detected object are not too far apart by a predefined number of tokens, and there are no conflicting/contradiction polarity terms in between.
2. For polarity terms in the form of preferences (e.g., product A is better than product B), the keyword before the preference term is assigned a polarity that is opposite to the polarity of the keyword after the preference term.
3. For Conjunction patterns (e.g., product A and product B are difficult to use), both keywords are assigned the same polarity.

*C. User Profile Extraction*

As mentioned, analysis results are generated against several user segments. These segments are generally based on gender, marital status, home locations and others. Twitter meta-data provides some of this information in a straightforward manner. But in other cases, users may decide not to add such information to their profile or they don't update in a timely manner. For example, a newly wedded person may not update his or her profile except after a while. Such implicit information may be detected through tweet text contents or the changes of profile description text. Back to the newly wedded example, the user may mention that he's going to get married. Another piece of information that isn't provided by twitter meta-data is parental status. In some cases it's of interest to know such information to understand the market segment. A user changing his or her profile with for example "Got a new a lovely child" would imply this person becoming a parent. Or a user tweeting "Facing motherhood challenges with my new baby" will imply that this person is a female and a parent. Due to space constraint, we cannot describe how user profiles were created in this paper.

*D. Results*

*1) Results from the annotated data set.*

The metrics we used to measure sentiment classification results are precision, recall and the standard F-measure. For each use case and sentiment polarity, 400 annotated tweets were randomly selected as the gold standard and then compared to the classification by the system. The classification results for Egyptian Teleco, Egyptian Government and Saudi Employment use cases are reported in Table III, IV and V, respectively.

For Egyptian Teleco, we obtained high precisions for the positive category (81.9%) and for the negative category (92.23%). The neutral category has a moderate precision (77.34%) but very high recall (91.43%). The F-scores range from 81.7% to 84.74%. The system was intentionally tuned towards high precision for the positive and negative categories for two reasons. First, positive and negative sentiments are much more interesting than neutral sentiment from a business perspective. Second, precision is often more important than recall in real-world applications. Incorrect classification of a false positive may have more significant negative business impact than a false negative.

TABLE III. CLASSIFICATION RESULTS FOR TWEETS IN EGYPTIAN TELECO USE CASE

| Classification | Precision | Recall | F-measure |
| --- | --- | --- | --- |
| Positive | 81.9 | 81.5 | 81.7 |
| Negative | 92.23 | 78.38 | 84.74 |
| Neutral | 77.34 | 91.43 | 83.8 |

TABLE IV. CLASSIFICATION RESULTS FOR TWEETS IN EGYPTIAN GOVERNMENT USE CASE

| Classification | Precision | Recall | F-measure |
| --- | --- | --- | --- |
| Positive | 81.25 | 83.20 | 82.21 |
| Negative | 90.46 | 77.45 | 83.45 |
| Neutral | 79.02 | 91.29 | 84.71 |

TABLE V. CLASSIFICATION RESULTS FOR TWEETS IN SAUDI EMPLOYMENT USE CASE

| Classification | Precision | Recall | F-measure |
| --- | --- | --- | --- |
| Positive | 52.23 | 67.25 | 58.8 |
| Negative | 77.92 | 71.48 | 74.56 |
| Neutral | 87.56 | 87.04 | 87.3 |

For Egyptian Government use case, we obtained high precision for the negative category (90.46%) and acceptable precisions for the positive and neutral categories (81.25% and 79.02% respectively). The F-scores range from 82.21% to 84.71%.

For Saudi Employment use case, the system performed moderately on the negative category (77.92% precision and 71.48% recall) and not so well on the positive category (52.23% precision and 67.25% recall). The system performed the best on the neutral category comparing to the other two categories. The F-scores for both positive and negative categories are the lowest among the three use cases. Reasons for such subpar performance are likely due to the sarcasm language used which makes it difficult to classify correctly, and also due to wrongly classifying tweets with news contents as positives, meanwhile they should be neutral, as tweets with news content do not carry user sentiment. As an example for sarcasm tweets that were wrongly classified as positives, consider

"#اجازه مع اني عاطلة بس احب ايام الاجازة؟؟ يارب يمددونها اسبوع زيادة"

Here, the user says that she like holidays although she is unemployed, and hope holiday to be extended! For another example, consider

"*تحيا العطالة يا جماعة ! . *وحدة بتلف الشوارع"

Also here the user salutes the unemployment.

In summary, our sentiment analysis has obtained satisfying results for the three use cases with F-scores ranging from 74.56% to 87.3% except the positive category in Saudi Employment. For the positive and negative categories that we are mostly interested in, the F-scores range from 74.56% to 84.74% except the positive category in Saudi Employment. The results are comparable to previous studies on sentiment analysis of dialectal Arabic despite of the many differences in test data sets and the approaches to sentiment classification [10, 12, 15].

*2) Results from the large Twitter data set.*

A 10% sample of entire Twitter traffic in January 2015 was collected and analyzed using the software system described in Section 3. Among the 1.05 billion tweets, 9158 of them were identified by the system as relevant to the Egyptian Teleco use case (i.e., containing keywords in a pre-defined set) and contains a positive or negative sentiment. Tweets that are relevant to the use case but have neutral or no sentiment are excluded.

This analysis examined Arabic tweets that are relevant to the Egyptian Teleco use case. At the end of the analysis, the system outputs a polarity label if either positive or negative sentiment is detected in a tweet. It also produced comprehensive author profiles including information like gender, marital status, job status, home location, and travel destinations. Enriching the sentiment results with this additional demographic and geographic information greatly helps examining and targeting a specific group of users in real business applications. In this use case of Teleco products and services, 6,536 positive and 2,622 negative sentiments were found. Among those tweets both male and female authors are more positive than negative towards the items we are interested in (Figure 1). We can also easily identify the countries (or even cities) with a large number of keywords of the Teleco products and services (Figure 2). Combining the location information and sentiment, it is instantly clear in which markets the users have a more favorable feedback and in which markets the user feedback are more balanced (Figure 3). With the capability of micro-segmentation, a company can selectively focus on the feedback from high-value users and geographic locations thus be more effective in carrying out marketing campaigns, improving products and services and etc.

## VII. CONCLUSION

We presented a system for classifying sentiment in Arabic text. The system is highly customizable and could be adapted to a new domain with moderate effort. It is capable of handling multiple sources of social media content, including Twitter, blogs and online discussion boards. Thanks the underlying technologies like Apache Hadoop, it can analyze massive amount of data in parallel as we demonstrated in the results. Unlike previous systems and models for Arabic, our system is tuned to high precision and moderate recall, which is often desirable in real world applications where sentiment analysis results have direct and real consequences in a business. Besides sentiment polarity, the system extracts from the data much more information, such as user profiles and geo-locations, which greatly extends the value of sentiment analysis. In summary, the system we described is a valuable and unique contribution to the Arabic sentiment analysis literature.

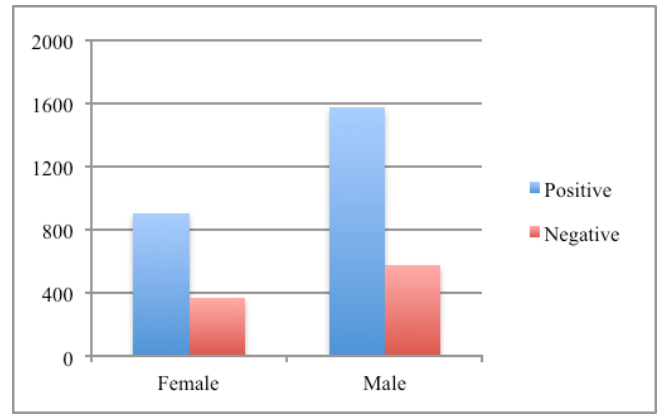

Figure 1. Number of tweets by sentiment polarity and gender of the author

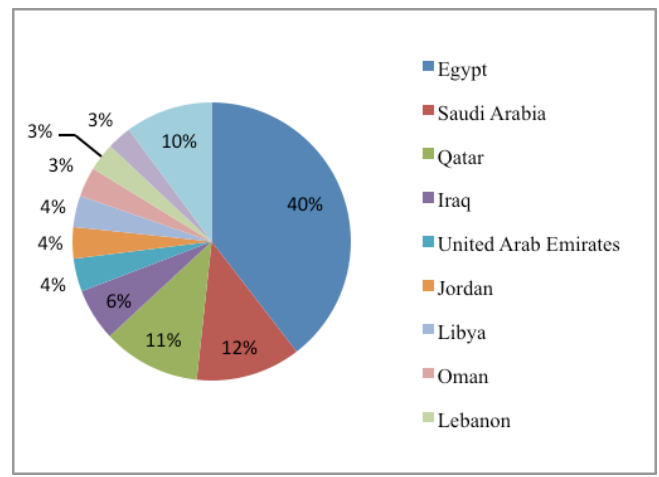

Figure 2. The originating countries of Arabic tweets with sentiment

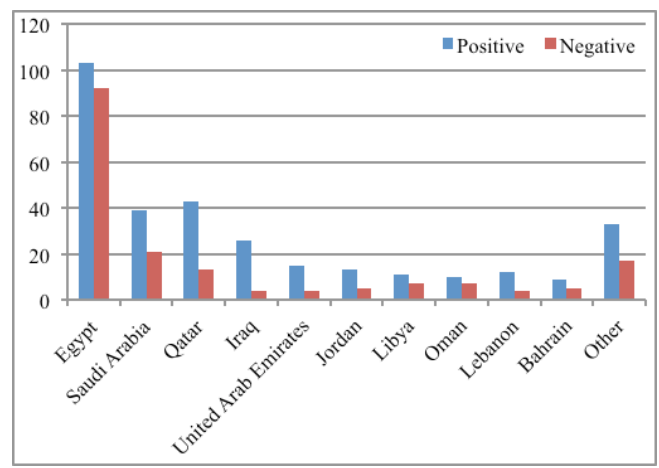

Figure 3. Number of tweets by sentiment polarity and originating country

For future work, the system can be improved in several aspects including enhancing the sentiment lexicons, fine-tuning the rules to capture languages patterns that are missed currently, detecting news content in text, and extending to more domains. The positive category in Saudi Employment use case is definitely a subject of further enhancement. Fortunately, in this system it can be done relatively easily by refining the sentiment lexicons (removing lexicon entries that caused excessive false positives) and adding more rules to capture the special and subtle language patterns used in this particular domain.